\documentclass{article}

\usepackage{chemformula} 
\usepackage[T1]{fontenc} 
\usepackage{amsmath, amssymb, graphicx, hyperref, float}
\usepackage{algorithm, algorithmic}
\usepackage{multirow}
\usepackage{booktabs}
\usepackage{float}
\usepackage{siunitx}
\usepackage{adjustbox}
\usepackage{tabularx}
\usepackage[a4paper, margin=3.0cm]{geometry}

\newcommand{\graycomment}[1]{\textcolor{gray}{// #1}}
\title{Rotational Sampling: \\A Plug-and-Play Encoder for Rotation-Invariant 3D Molecular GNNs}
\author{Dian Jin \\ \small The Hong Kong Polytechnic University, Hong Kong SAR, China}
\date{}

\begin{document}
\maketitle
\begin{abstract}

Graph neural networks (GNNs) have achieved remarkable success in molecular property prediction. However, traditional graph representations struggle to effectively encode the inherent 3D spatial structures of molecules, as molecular orientations in 3D space introduce significant variability, severely limiting model generalization and robustness. Existing approaches primarily focus on rotation-invariant and rotation-equivariant methods. Invariant methods often rely heavily on prior knowledge and lack sufficient generalizability, while equivariant methods suffer from high computational costs.
To address these limitations, this paper proposes a novel plug-and-play 3D encoding module leveraging rotational sampling. By computing the expectation over the SO(3) rotational group, the method naturally achieves approximate rotational invariance. Furthermore, by introducing a carefully designed post-alignment strategy, strict invariance can be achieved without compromising performance.
Experimental evaluations on the QM9 and C10 Datasets demonstrate superior predictive accuracy, robustness, and generalization performance compared to existing methods. Moreover, the proposed approach maintains low computational complexity and enhanced interpretability, providing a promising direction for efficient and effective handling of 3D molecular information in drug discovery and material design.

\end{abstract}

\noindent\textbf{Keywords:} Graph neural networks; Graph representation learning; Molecule property prediction

\section{Introduction}

The discovery and development of novel drugs  and materials traditionally requires extensive experimental efforts, consuming significant time and resources\cite{vamathevan2019applications,oganov2019structure}. Accelerating these processes is crucial for addressing pressing global challenges, from developing treatments for public health menace posed by antimicrobial resistance \cite{miethke2021towards} to creating sustainable, resilient, clean material solutions \cite{stier2024materials}.

Graph Neural Networks (GNNs) have revolutionized various fields, from social network analysis\cite{li2023survey} to e-commerce product recommendation\cite{gao2023survey}, by effectively modeling and analyzing relational data. More recently, their application in biochemistry has led to significant breakthroughs, particularly in molecular property prediction. Accurate molecular property prediction plays a critical role in drug discovery\cite{fang2022geometry}, materials design\cite{fung2021benchmarking}, and other scientific advancements\cite{wang2020graph,anandhi2024systematic}. 

However, encoding molecular structures as graphs presents unique challenges, primarily due to the lack of inherent 3D spatial information in standard graph representations.
In molecular systems, atoms are typically modeled as nodes in a graph, and their 3D positional information is crucial for understanding molecular properties. Researchers typically rely on Cartesian coordinates to represent molecular structures in Euclidean space, but these coordinates are inherently rotation-dependent. A single molecule can have infinitely many valid orientations in 3D space, introducing noise and uncertainty into models and hindering their ability to learn stable and generalizable positional embeddings \cite{rao2022quantitative, zhu2022hignn}.
Addressing this challenge is particularly important for improving the prediction accuracy and robustness of molecular property prediction models\cite{fang2022geometry}.

To overcome this limitation, methods for embedding 3D information into GNNs have been developed\cite{fei2024rotation}:

First, rotation-Invariant Methods. These methods aim to encode 3D structural information in a way that is independent of molecular orientation. Currently, rotation-invariant methods for 3D molecular embeddings loosely fall into two main categories:
\begin{enumerate}
  \item Feature Engineering–Based Methods: These techniques exploit handcrafted descriptors grounded in chemical intuition. For instance, SchNet\cite{schutt2018schnet} employs a radial basis function (RBF) expansion of interatomic distances, while SphereNet\cite{coors2018spherenet} and GEM\cite{fang2022geometry} integrate angular features. Despite their success, these methods are constrained by their dependence on prior knowledge and often struggle to generalize.
  \item Alignment–Based Methods: These approaches enforce rotation invariance by registering each molecule to a canonical frame, e.g., via principal component analysis\cite{pop2023rotation}. Although conceptually simple, they tend to limit the expressive power of learned representations. More elaborate alignment schemes have been proposed\cite{NEURIPS2023_fb4a7e35}, but they still require user‐defined anchors and coordinate constructions, which introduce inevitable biases and restrictions.
\end{enumerate}

Second, rotation-Equivariant Methods.
These methods maintain equivariant representations under 3D rotations by constructing operations that preserve equivariance throughout the network. Invariant features most commonly derived by taking the norms of equivariant tensors, can subsequently be used for downstream tasks. Nevertheless, ensuring strict equivariance often leads to a significant increase in computational cost\cite{passaroReducingSO3Convolutions2023}. Examples include EGNN\cite{satorrasEquivariantGraphNeural2022}, Cormorant\cite{andersonCormorantCovariantMolecular2019}, and the SE(3)-Transformer\cite{fuchsSE3Transformers3DRotoTranslation2020}.

The above methods are either constrained by data or face limitations in fitting capabilities. Therefore, to enhance the model's fitting ability without requiring complex calculations, a convolution-based method was proposed \cite{zhangPointGATQuantumChemical2024}. This method achieves state-of-the-art performance on multiple datasets, with the 3D embedding module having relatively low computational complexity. However, it does not consider rotation invariance. Thus, despite these advancements, existing methods face significant challenges in balancing rotation invariance, generalizability, and predictive accuracy. 

In addition, in many molecule-related fields, data acquisition is challenging. On one hand, obtaining experimental data can be expensive or time-consuming. For example, the research \cite{leePrincipalOdorMap2023} utilizes the graph representation of molecules to predict their odors. However, annotating odors requires significant resources (the smell of molecules cannot be simulated and requires gathering personnel for subjective evaluation), and thus the mentioned dataset contains only 5k samples. On the other hand, some fields inherently lack large amounts of known data. For example, in the study of drug targets, the number of FDA-approved drug targets is only 893, of which only 667 are targets for human diseases~\cite{santos2017comprehensive}. This indicates that despite the continuous progress in drug development, the number of known drug targets remains extremely limited.
Therefore, it is crucial to build effective models with limited data.

To address these gaps, we propose a plug-and-play module that enhances rotation-invariant representations and demonstrates excellent performance on small datasets. Unlike feature engineering or alignment methods, our approach extracts information beyond manually defined features, enabling more robust and generalizable embeddings.
We validate our method on the QM9 Dataset and C10 Dataset, demonstrating its superior performance, generalization ability, and versatility compared to existing approaches. 
Our main contributions are listed below:

\begin{enumerate}
    \item We propose a novel 3D information encoding module based on the expectation over the SO(3) group. This module enhances the rotational invariance and fitting capability of molecular graph neural networks. 
    \item We demonstrate that, under identical base model settings, integrating our method with engineered geometric features for 3D encoding yields the best performance, outperforming solely engineered features.
    \item We conduct extensive experiments and ablation studies on subsets of the QM9 Dataset of different sizes to validate the robustness, generalization, and versatility of our approach. Our analysis suggests that predefined alignment is detrimental, and on small datasets, our method demonstrates stronger feature extraction capabilities.
\end{enumerate}

\section{Methodology}

\subsection{Preliminaries of Graph Neural Networks}

GNN has emerged as a powerful architecture for processing data with complex relational structures. Since molecules can be represented as graphs, with atoms as nodes and bonds as edges, GNN has become increasingly important in molecular-related analyses \cite{wu2023chemistry}.

GNN learns representations by recursively aggregating and transforming feature information from neighboring nodes. In practice, Message Passing Neural Network (MPNN) is a popular GNN architecture \cite{gilmer2017neural}. MPNN formalizes the process of information exchange between nodes through two phases: message passing and readout. More specifically, for a given graph $G: = (V, E)$ with initialized feature representations $\mathbf{h}_v^{0}$ for each node $v \in V$ and predefined embeddings $\mathbf{e}_{uv}$ for each edge $(u, v) \in E$, MPNN updates these features iteratively in the message passing phase by the following update rules:

\begin{equation}
\label{eq:update_m_MPNN}
\mathbf{m}_v^{l+1} = \sum_{w \in \mathcal{N}(v)} M_l(\mathbf{h}_v^l, \mathbf{h}_w^l, \mathbf{e}_{vw}) \quad \forall v \in V,
\end{equation}

and

\begin{equation}
\mathbf{h}_v^{l+1} = U_l(\mathbf{h}_v^l, \mathbf{m}_v^{l+1}) \quad \forall v \in V,
\end{equation}

where $\mathbf{h}_v^l$ is the feature representation of node $v$ at iteration $l$, $M_l$ and $U_l$ are the message passing function and the update function at iteration $l$, respectively.








After obtaining the representations $\mathbf{h}^{L}_v$ for all $v \in V$ from the message passing phase at the final iteration $L$, the readout phase then computes a graph-level representation by aggregating the final node representations:

\begin{equation}
\hat{\mathbf{y}} = \phi(\mathbf{h}_v^L ~|~ v \in V),
\end{equation}
where $\phi$ is the readout function, typically implemented as a summation, average, or more sophisticated pooling mechanism in practice. We will then apply the learned feature representation $\hat{\mathbf{y}}$ of the graph in the downstream tasks. 



\subsection{Invariant Representation Learning}

The GNN mentioned above has not yet considered the 3D information of nodes. Although directly concatenating their coordinates to $\mathbf{h}_v^0$ is a straightforward idea, the initial states of the molecular orientations are random. Therefore, this straightforward idea cannot be effectively utilized by GNNs. To address this challenge,  several feature extraction methods have been proposed, typically including distances and angles, which possess inherent rotational invariance. However, these methods usually lack strong generalization capabilities and cannot distinguish chiral structures of molecules
, which limits their applications. Recently, a PointNet-based approach called PointGAT \cite{zhangPointGATQuantumChemical2024} has been proposed, which can help to distinguish chiral structures. However, it may break the rotational invariance of the learned representation, which is an important property required by many downstream applications. In this paper, we propose a method to better address the challenges in distinguishing chiral structures of molecules and learning rotational invariant representations. 


\subsubsection{Problem Setup}
Given an input graph $G = (V, E, \mathbf{X})$ where $\mathbf{X} \in \mathbb{R}^{|V| \times 3}$ denotes the spatial coordinates, where each node $ v \in V $ is associated with a 3D coordinate $ \mathbf{x}_v $. The goal is to ensure that the output of the GNN remains invariant under arbitrary 3D rotations of the input.

To mathematically define this problem, we consider any rotation matrix $ \mathbf{R} \in SO(3) $. For a given set of model parameters $ \theta $, the output of the GNN $f_{\theta}$  for the rotated input $ \mathbf{R}\mathbf{x}_v $ should satisfy:
$$
f_{\theta}(\mathbf{R}\mathbf{x}_v) = f_{\theta}(\mathbf{x}_v)
$$
ensuring rotational invariance.

\subsubsection{Theoretical Framework}

We propose an objective function to address this issue. This objective function is designed by constructing the minimum of expected loss as follows:

\begin{equation}
    \min_\theta  \left[ \ell\left(\mathbb{E}_{\mathbf{R} \in SO(3)}f_{\theta}(Rx), \mathbf{y}\right) \right]
    \label{eq:minE_Inv}
 \end{equation}

\begin{equation}
    \min_\theta \mathbb{E}_{\mathbf{R} \in SO(3)} \left[ \ell\left(\theta, \mathbf{R}\mathbf{x}, \mathbf{y}\right) \right]
    \label{eq:minE}
 \end{equation}

where $ \ell(\cdot) $ is a task-specific loss function (e.g., regression or classification loss), 
$ \mathbf{R} \in SO(3) $ represents a rotation matrix sampled from the group of 3D rotations,
$ \mathbf{x} $ and $ \mathbf{y} $ are the input and target outputs, respectively, and
$ \theta $ are the learnable parameters of the network.
The network aims to produce consistent outputs across all possible 3D rotations.


\paragraph{Definition of $ f_\theta $}

The learned function, or 3D encoder $ f_\theta $ is implicitly defined by the optimization problem through the expected loss. For a given input $ \mathbf{x} $, the output of $ f_\theta $ depends on the expectation of the predictions over all rotated versions of $ \mathbf{x} $. Mathematically, for a fixed parameter $ \theta $, we can express $ f_\theta(\mathbf{x}) $ as:

$$
f_\theta(\mathbf{x}) = \mathbb{E}_{\mathbf{R} \in SO(3)} \left[ g_\theta(\mathbf{R}\mathbf{x}) \right]
$$

where $ g_\theta $ represents the raw output of the model. The expectation ensures that $ f_\theta(\mathbf{x}) $ depends not on a single orientation of $ \mathbf{x} $, but on the aggregate behavior of $ g_\theta $ across all $ \mathbf{R} \in SO(3) $.

\paragraph{Rotational Invariance}

To prove that $ f_\theta $ is rotationally invariant, we need to show that for any $ \mathbf{R}_0 \in SO(3) $:

$$
f_\theta(\mathbf{R}_0\mathbf{x}) = f_\theta(\mathbf{x})
$$

By definition of $ f_\theta $, we have:

$$
f_\theta(\mathbf{R}_0\mathbf{x}) = \mathbb{E}_{\mathbf{R} \in SO(3)} \left[ g_\theta(\mathbf{R}(\mathbf{R}_0\mathbf{x})) \right].
$$

Using the invariance property of the Haar measure, which states that for any fixed $ \mathbf{R}_0 \in SO(3) $ and any measurable function $ h: SO(3) \to \mathbb{R} $,

$$
\int_{SO(3)} h(\mathbf{R} \mathbf{R}_0) \, d\mu(\mathbf{R}) = \int_{SO(3)} h(\mathbf{R}) \, d\mu(\mathbf{R}),
$$

we can rewrite $ f_\theta(\mathbf{R}_0\mathbf{x}) $ as:

$$
f_\theta(\mathbf{R}_0\mathbf{x}) = \mathbb{E}_{\mathbf{R} \in SO(3)} \left[ g_\theta(\mathbf{R}\mathbf{x}) \right].
$$

This is exactly the definition of $ f_\theta(\mathbf{x}) $. Therefore, we have:

$$
f_\theta(\mathbf{R}_0\mathbf{x}) = f_\theta(\mathbf{x}),
$$

which proves that $ f_\theta $ is rotationally invariant.

\subsubsection{Practical Implementation}

While infinite sampling is theoretically optimal, practical constraints necessitate a finite approximation. Specifically, we sample a finite set of rotations $ \{{\mathbf{R}}_i\}_{i=1}^N $ and approximate $ f_\theta(\mathbf{x}) $ as:

$$
f_\theta(\mathbf{x}) \approx \frac{1}{N} \sum_{i=1}^N g_\theta({\mathbf{R}}_i \mathbf{x})
$$

More specifically, the Sampling and Aggregation Framework comprises three principal steps:

\paragraph{Rotation Sampling}
Given $G = (V, E, \mathbf{X})$, we construct a set of rotation matrices $\{\mathbf{R}_i\}_{i=1}^N$ sampled from $SO(3)$ group. These rotations are generated through a quasi-uniform sampling strategy to ensure comprehensive coverage of the rotation space. For each node $v \in V$, we obtain a collection of rotated coordinates $\{\mathbf{R}_i\mathbf{x}_v\}_{i=1}^N$.

\paragraph{Neural Processing Pipeline}
The framework employs parallel processing of the rotated variants through function $g_\theta$. Each rotated instance generates an independent feature representation:

$$
\mathbf{y}_i = g_\theta(G, \mathbf{R}_i\mathbf{X}), \quad i \in \{1,\ldots,N\}
$$

\paragraph{Statistical Aggregation Mechanism}
The final representation is obtained through a statistical aggregation operator $\mathcal{A}$:

$$
\mathbf{y} = \mathcal{A}(\{\mathbf{y}_i\}_{i=1}^N)
$$

This formulation asymptotically approaches rotational invariance as $N \to \infty$, consistent with our theoretical analysis. 

\subsection{Alignment Strategy}

Given the set of atom coordinates $\mathbf{X} \in \mathbb{R}^{|V| \times 3}$ (as mentioned in the previous section) representing molecular coordinates, a Principal Component Analysis (PCA)-based alignment strategy enforces rotation-invariant encoding and guarantees that any rotationally equivalent molecular configuration will produce identical graph representations through the canonical projection $\mathbf{\tilde{X}}$, while preserving the intrinsic geometric relationships through proper handedness maintenance and sign consistency enforcement.  See Algorithm \ref{alg:pca} for the pseudocode.

\begin{algorithm}[ht]
\caption{Point Cloud Alignment via Principal Component Analysis}
\label{alg:pca}
\begin{algorithmic}[1]
\REQUIRE Point cloud $\mathbf{X} \in \mathbb{R}^{|V| \times 3}$
\ENSURE Canonical projection $\mathbf{\tilde{X}}$

\STATE \graycomment{Compute the principal component vectors}
\STATE $\mathbf{P} \gets \text{PCA}(\mathbf{X})$ \graycomment{$\mathbf{P}$ is a $3 \times 3$ matrix}

\STATE \graycomment{Ensure right-handed coordinate system}
\IF{$\det(\mathbf{P}) < 0$}
    \STATE $\mathbf{P}_{3,:} \gets -\mathbf{P}_{3,:}$ 
\ENDIF

\STATE \graycomment{Project point cloud onto principal components}
\STATE $\mathbf{\tilde{X}} \gets \mathbf{X} \mathbf{P}^\top$

\STATE \graycomment{Normalize orientation to ensure consistent quadrant placement}
\STATE $\mathbf{p_{ref}} \gets \mathbf{\tilde{X}}[\arg\max(\alpha(\mathbf{\tilde{X}})), :]$ \graycomment{$\alpha$ counts the non-zero for each row}
\STATE \graycomment{\{O\} and \{Z\} are sequence, $|\{O\}|+|\{Z\}| = 3$ }
\STATE $\{O\} = \{ o \mid \mathbf{p_{ref}}[o] = 0, o=1,2,3\}$ \graycomment{$o_i$ for elements in $\{O\}$}
\STATE $\{Z\} = \{ z \mid \mathbf{p_{ref}}[z] \neq 0, z=1,2,3 \}$ \graycomment{$z_i$ for elements in $\{Z\}$}

\STATE  $c\gets 0$ \graycomment{Count flip number}
\FOR{$z$ in $\{Z\}$}
\IF{$\mathbf{p_{ref}}[z] < 0$}
    \STATE $c \gets c+ 1$
\ENDIF
\ENDFOR

\STATE  \graycomment{Ensure right-handed}
\IF{$c \equiv 1 \mod 2$}
    \IF{$|\{Z\}| < 3$}
        \STATE $\mathbf{\tilde{X}}[:,o_1] \gets -\mathbf{\tilde{X}}[:,o_1]$ 
    \ELSE
            \STATE $\mathbf{\tilde{X}}[:,3] \gets -\mathbf{\tilde{X}}[:,3]$ 
    \ENDIF
\ENDIF

\RETURN $\mathbf{\tilde{X}}$
\end{algorithmic}
\end{algorithm}

\subsection{Model Architecture}

\subsubsection{3‑D Geometric Encoder}

The point cloud $\mathbf{X} \in \mathbb{R}^{n \times 3}$ contains the Cartesian coordinates of the $n$ atoms in a molecule.
Then we sample $k$ rotation matrices
${\mathbf{R}^{(j)}} \subset \mathrm{SO}(3), j = 1, \dots, k$ and generate rotated views:
\begin{equation}
\mathbf{X}^{(j)} = \mathbf{X} \mathbf{R}^{(j)\top},
\qquad j = 1, \dots, k .
\end{equation}
Optionally, atomic embedding information can be included by concatenating the rotated coordinates with atom-wise embeddings, i.e.,
$\mathcal{X}^{(j)}_0 = [\mathbf{X}^{(j)}\| \mathrm{Emb}(z)]$,
where $\|$ denotes concatenation and $\mathrm{Emb}(z)$ denotes the embedding of atoms.
In this case, the feature dimension becomes $l \geq 3$.

\paragraph{Point‑wise convolution.}
We apply $\tau$ successive $1{\times}1$ convolutions to $\mathcal{X}^{(j)}_0$ to extract local geometric features.
\begin{equation}
  \mathcal{X}^{(j)}_{\ell}=
  \sigma (\mathrm{BN}(
    \mathrm{Conv}(\mathcal{X}^{(j)}_{\ell-1}))
  ),
  \quad \ell=1,...,\tau
  \label{eq:pw_conv}
\end{equation}
where $\mathbf{W}^{(\ell)}$ are learnable parameters for layer~$\ell$,  BN is the batch Norm, 
and $\sigma(\cdot)$ is the  activation function.  

\paragraph{Global pooling.}
As the output should be invariant to the ordering of atoms within the list (permutation invariance), a pooling operation is employed to aggregate atomic features into a unified view fingerprint.
\begin{equation}
  \mathbf{p}^{(j)}=
  \mathrm{Pool}(\mathcal{X}^{(j)}_{\tau})
  \in\mathbb{R}^{d_{p}},
  \label{eq:view_fp}
\end{equation}
where $d_{p}$ is the fingerprint length.

\paragraph{Multi‑view fusion.}
Averaging over the $k$ views yields a rotation‑robust representation:
\begin{equation}
  \mathbf{p}=\frac{1}{k}\sum_{j=1}^{k}\mathbf{p}^{(j)}, 
  \label{eq:avg_views}
\end{equation}
which approximates an integral over $\mathrm{SO}(3)$.

\subsubsection{Feature Fusion and \texorpdfstring{$\ell_{1}$}{L1} Regularisation}\label{subsec:fusion}

The geometric vector $\mathbf{p}$ is concatenated with 
the global representation $\mathbf{g}$ to form
\begin{equation}
  \mathbf{u}=
  [\mathbf{g}\,\|\,\mathbf{p}]
  \in\mathbb{R}^{d_{u}} ,
  \label{eq:concat}
\end{equation}
where $d_{u}$ is the total feature dimension.

\subsubsection{Regression Head}

A multilayer perceptron  $\gamma$ maps $\mathbf{u}$ to predictions:
\begin{align}
  \hat{\mathbf{y}} &= \gamma(\mathbf{u}),
\end{align}

The training objective combines task loss and sparsity:
\begin{equation}
  \mathcal{L}=
  \mathcal{L}_y+\lambda\lVert\mathbf{u}\rVert_{1},
  \label{eq:loss_total}
\end{equation}
where $\mathcal{L}_y$ is the prediction loss,  
and $\lambda$ controls the strength of the $\ell_{1}$ penalty.






\section{Experiments}
\subsection{Experimental Setup}

To evaluate the performance of our proposed method, we conducted experiments on two datasets: QM9 and C10. The experiments are designed to highlight the advantages of our approach in capturing 3D molecular features, particularly on small-scale datasets. All experiments are based on a consistent baseline model architecture to ensure fair comparisons. Specific implementation details and dataset descriptions are provided below.

\subsection{Datasets}

\subsubsection{QM9 Dataset}

The QM9 Dataset \cite{ramakrishnan2014quantum} is a widely used benchmark for quantum machine learning. To emphasize the performance of our model on small-scale datasets, we randomly sampled a subset from QM9, consisting of 3k and 10k molecules. The data was sourced from the DGL Graph Library. Moreover, we also conducted tests on the entire dataset (130k).

The QM9 Dataset contains molecular structures with associated quantum properties, providing an ideal testbed for evaluating GNN that incorporate 3D spatial information.

\subsubsection{C10 Dataset}

The C10 Dataset \cite{zhangPointGATQuantumChemical2024} contains 11,841 monoterpene carbocation intermediates with diverse 3D geometries and high-quality quantum-chemically calculated energies. It covers 75 distinct skeletons, including monocyclic and bicyclic structures with varying ring sizes (e.g., five-membered rings), and retains detailed 3D spatial configurations. While the carbocations share similar 2D structures, their 3D conformations and energies vary significantly, posing a challenge for 3D-sensitive GNNs. This dataset’s complex chemical space and accurate energy labels make it ideal for evaluating GNN performance in capturing 3D molecular features.

We performed a random split of the dataset into training, validation, and test sets. This dataset was specifically chosen to evaluate the ability of our method to encode complex 3D molecular features.

\subsection{Model and Methods}

\subsubsection{Backbone Model}

Our backbone model architecture consists of the following components:

\begin{itemize}
    \item Graph Embedding: A Graph Neural Network (GNN) is used to encode graph structures. For the QM9 Dataset, we use a simple MPNN, while for the C10 Dataset, we use AttentiveFP (for comparison with SOTA methods).
    \item 3D Encoder: Different methods are employed for comparative analysis, including our proposed method.
    \item Regression Head: An MLP is used to predict numerical outputs for the regression task.
\end{itemize}

\begin{figure}[ht]
    \centering
    \includegraphics[width=0.8\textwidth]{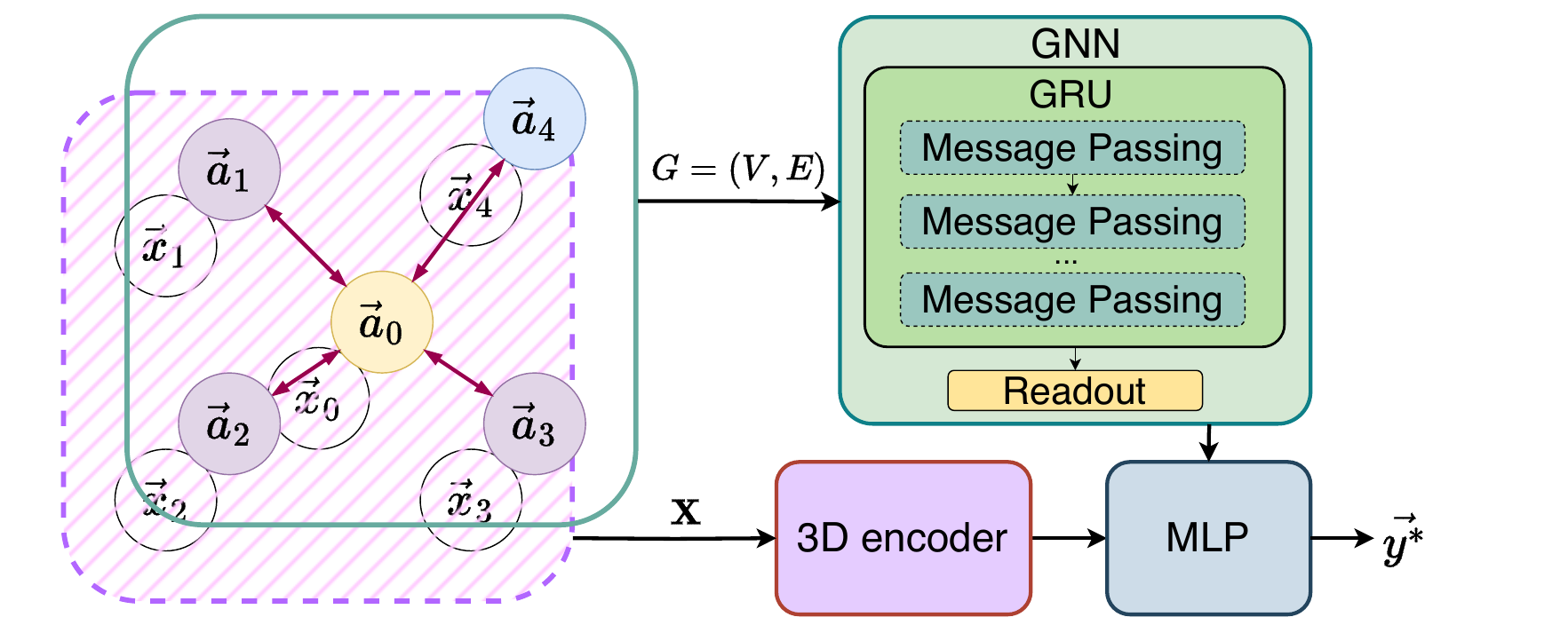} 
    \caption{Overview of the model pipeline to gather 3D information and molecule information.}
    \label{fig:gnn}
\end{figure}

The overall pipeline is illustrated in Fig.~\ref{fig:gnn}. All methods are based on this backbone model to ensure fairness during comparisons.

\subsubsection{Training Details}

For the QM9 Dataset (3k and 10k subsets), the model was trained for 800 epochs with a batch size of 128 and a learning rate of 0.001, and five-fold cross-validation was employed.
For the 130k dataset, to ensure training stability, we first performed 5-fold cross-validation on a randomly sampled 13k subset to select parameters. Then, using the complete dataset, we randomly selected 70\% as the training set and the remainder as the test set, training for 800 epochs with a batch size of 256.
We utilized the AdamW optimizer and MSELoss as the loss function. The model parameters included a uniform 32-dimensional embedding for atomic features and a 128-dimensional embedding for 3D molecular information. Models are trained on one Nvidia A100.

For the C10 Dataset, the data was split into training, validation, and test sets by original work \cite{zhangPointGATQuantumChemical2024}. The state-of-the-art model, PointGAT, which combines AttentiveFP with PointNet, served as a benchmark. To ensure a fair comparison, we also used AttentiveFP as backbone GNN.

\section{Results}

\subsection{Performance on QM9 Dataset}
The performance of several models on QM9 Dataset is shown in Table \ref{tab:3k}, \ref{tab:10k}, and \ref{tab:130k}.
The results indicate that the proposed method demonstrates significant performance advantages across different tasks and data scales. On most tasks, it outperforms traditional approaches that incorporate engineered features such as RBF (for distance embedding) or AGL (for angle embedding), achieving lower mean absolute error (MAE). This outcome validates the robustness and generalization capability of our method, suggesting that it can effectively capture key molecular properties without relying on manually designed prior knowledge. 

This highlights that while engineered features (RBF and AGL) can capture certain useful information about molecular structures, they have inherent limitations in representational capacity. Relying solely on engineered features is insufficient for fully characterizing molecular properties. In contrast, our method adaptively learns the intricate internal relationships within molecules, compensating for the deficiencies of engineered features and enabling more accurate predictions. Furthermore, when combined with engineered features (e.g., Ours + RBF + AGL), performance is further improved, achieving state-of-the-art results on many tasks.

\newcommand{\tscale}{0.95}

\begin{table}[ht]
\centering
\caption{MAE Results of indicators on QM9 Dataset (3k)}
\begin{tabularx}{\tscale\textwidth}{l*{6}{>{\centering\arraybackslash}X}}
\toprule
Methods & A & B & C & Cv & G$_\text{atom}$ & G \\
\midrule
MPNN & 0.7051 & 0.2216 & 0.1448 & 1.4662 & 2.6000 & 375.5999 \\
+RBF+AGL & 0.5579 & 0.1468 & 0.0905 & 0.7717 & 1.3187 & 188.0317 \\
+PointNet & 0.5634 & 0.1436 & 0.0799 & 0.8230 & 1.2845 & 197.9472 \\
\midrule
Ours(pre-align) & 0.4877 & 0.1243 & 0.0724 & 0.7094 & 1.0921 & 175.8335 \\
Ours(post-align) & 0.4504 & 0.1249 & 0.0733 & 0.6889 & 1.0901 & 167.9127 \\
Ours & 0.4639 & 0.1250 & 0.0722 & 0.6737 & 0.9953 & 160.5483 \\
Ours(+RBF+AGL) & \textbf{0.4461} & \textbf{0.1224} & \textbf{0.0682} & \textbf{0.5470} & \textbf{0.8584} & \textbf{131.5533} \\
\bottomrule
\end{tabularx}

\vspace{0.2cm}

\begin{tabularx}{\tscale\textwidth}{l*{6}{>{\centering\arraybackslash}X}}
\toprule
Methods & H$_\text{atom}$ & H & U0$_\text{atom}$ & U0 & U$_\text{atom}$ & U \\
\midrule
MPNN & 2.8405 & 375.6422 & 2.8053 & 375.4827 & 2.8398 & 375.9985 \\
+RBF+AGL & 1.4315 & 187.8804 & 1.4124 & 187.6919 & 1.4237 & 188.2252 \\
+PointNet & 1.3845 & 198.7527 & 1.3672 & 197.8020 & 1.3729 & 199.3016 \\
\midrule
Ours(pre-align) & 1.1755 & 176.7581 & 1.1524 & 176.0158 & 1.1723 & 175.5180 \\
Ours(post-align) & 1.1606 & 168.3277 & 1.1474 & 168.4175 & 1.1641 & 166.7949 \\
Ours & 1.0658 & 160.4646 & 1.0573 & 160.1102 & 1.0659 & 160.5302 \\
Ours(+RBF+AGL) & \textbf{0.9155} & \textbf{132.0798} & \textbf{0.9076} & \textbf{131.4517} & \textbf{0.9117} & \textbf{132.2870} \\
\bottomrule
\end{tabularx}

\vspace{0.2cm}

\begin{tabularx}{\tscale\textwidth}{l*{7}{>{\centering\arraybackslash}X}}
\toprule
Methods & alpha & gap & homo & lumo & mu & r2 & zpve \\
\midrule
MPNN & 3.0439 & 0.3660 & 0.2366 & 0.3088 & 0.7867 & 113.9440 & 0.1955 \\
+RBF+AGL & 1.6782 & 0.2928 & 0.1951 & 0.2321 & 0.7354 & 68.6023 & 0.1151 \\
+PointNet & 1.5628 & 0.3185 & 0.2076 & 0.2544 & 0.7297 & 56.5184 & 0.1142 \\
\midrule
Ours(pre-align) & 1.3355 & 0.2964 & 0.1961 & 0.2429 & 0.7147 & 48.8575 & 0.1007 \\
Ours(post-align) & 1.3096 & 0.3025 & 0.2043 & 0.2451 & 0.7057 & 49.5400 & 0.1018 \\
Ours & 1.3051 & 0.2976 & 0.1949 & 0.2499 & 0.7098 & 48.0871 & 0.0967 \\
Ours(+RBF+AGL) & \textbf{1.2176} & \textbf{0.2783} & \textbf{0.1881} & \textbf{0.2267} & \textbf{0.7039} & \textbf{46.5014} & \textbf{0.0952} \\
\bottomrule
\end{tabularx}

\vspace{0.2cm}

\label{tab:3k}
\end{table}

\begin{table}[H]
\centering
\caption{MAE Results of indicators on QM9 Dataset (10k)}
\begin{tabularx}{\tscale\textwidth}{l*{6}{>{\centering\arraybackslash}X}}
\toprule
Methods & A & B & C & Cv & G$_\text{atom}$ & G \\
\midrule
+RBF+AGL & 0.3767 & 0.1016 & 0.0619 & 0.4417 & 0.6956 & 91.3300 \\
Ours(+RBF+AGL) & \textbf{0.3196} & \textbf{0.0936} & \textbf{0.0538} & \textbf{0.3890} & \textbf{0.5793} & \textbf{73.8734} \\
\bottomrule
\end{tabularx}

\vspace{0.2cm}

\begin{tabularx}{\tscale\textwidth}{l*{6}{>{\centering\arraybackslash}X}}
\toprule
Methods & H$_\text{atom}$ & H & U0$_\text{atom}$ & U0 & U$_\text{atom}$ & U \\
\midrule
+RBF+AGL & 0.7472 & 91.2984 & 0.7386 & 91.3048 & 0.7433 & 91.3027 \\
Ours(+RBF+AGL) & \textbf{0.6116} & \textbf{73.8714} & \textbf{0.6056} & \textbf{73.8725} & \textbf{0.6091} & \textbf{73.8714} \\
\bottomrule
\end{tabularx}

\vspace{0.2cm}

\begin{tabularx}{\tscale\textwidth}{l*{7}{>{\centering\arraybackslash}X}}
\toprule
Methods & alpha & gap & homo & lumo & mu & r2 & zpve \\
\midrule
+RBF+AGL & 0.9925 & 0.2240 & 0.1514 & 0.1779 & \textbf{0.6211} & 41.4815 & \textbf{0.0662} \\
Ours(+RBF+AGL) & \textbf{0.9046} & \textbf{0.2221} & \textbf{0.1505} & \textbf{0.1750} & 0.6254 & \textbf{32.3087} & 0.0684 \\
\bottomrule
\end{tabularx}

\vspace{0.2cm}

\label{tab:10k}
\end{table}

\begin{table}[ht]
\centering
\caption{MAE Results of indicators on QM9 Dataset (130k, full dataset)}
\begin{tabularx}{\tscale\textwidth}{l*{6}{>{\centering\arraybackslash}X}}
\toprule
Methods & A & B & C & Cv & G$_\text{atom}$ & G \\
\midrule
+RBF+AGL & \textbf{21.3499} & 0.1044 & \textbf{0.0577} & \textbf{0.2430} & 0.3334 & 39.0774 \\
Ours(+RBF+AGL) & 21.8241 & \textbf{0.1041} & 0.0584 & 0.2480 & \textbf{0.3025} & \textbf{33.1646} \\
\bottomrule
\end{tabularx}

\vspace{0.2cm}

\begin{tabularx}{\tscale\textwidth}{l*{6}{>{\centering\arraybackslash}X}}
\toprule
Methods & H$_\text{atom}$ & H & U0$_\text{atom}$ & U0 & U$_\text{atom}$ & U \\
\midrule
+RBF+AGL & 0.3597 & 39.0752 & 0.3550 & 39.0762 & 0.3577 & 39.0781 \\
Ours(+RBF+AGL) & \textbf{0.3264} & \textbf{33.1618} & \textbf{0.3218} & \textbf{33.1597} & \textbf{0.3244} & \textbf{33.1625} \\
\bottomrule
\end{tabularx}

\vspace{0.2cm}

\begin{tabularx}{\tscale\textwidth}{l*{7}{>{\centering\arraybackslash}X}}
\toprule
Methods & alpha & gap & homo & lumo & mu & r2 & zpve \\
\midrule
+RBF+AGL & 0.5822 & 0.1541 & 0.1078 & \textbf{0.1213} & 0.4694 & 31.3173 & 0.0392 \\
Ours(+RBF+AGL) & \textbf{0.5700} & \textbf{0.1518} & \textbf{0.1042} & 0.1222 & \textbf{0.4572} & \textbf{29.9671} & \textbf{0.0371} \\
\bottomrule
\end{tabularx}

\vspace{0.2cm}

\label{tab:130k}
\end{table}

The PointGAT method (+PointNet) is based on PointNet to achieve 3D embeddings but does not account for the impact of rotational invariance. This means that the same input may lead to unstable outputs due to the randomness of the initial state. 
Our proposed method enhances rotational invariance by calculating the expectation under SO(3). The average and maximum results from multiple experiments are shown in Table \ref{tab:ri}, demonstrating that our method produces more stable outputs.

\begin{table}[ht]
\centering
\caption{Rotational invariance error}
\begin{tabular}{lSS}
\toprule
 Methods    &    {Mean} &     {Max} \\
\midrule
 PointNet   & 15.3920 & 22.1453 \\
 Our Method &  6.9289 & 8.8069 \\
 +align     &  0.0000 &  0.0000 \\
\bottomrule
\end{tabular}
\label{tab:ri}
\end{table}

We also explored a fixed molecular alignment strategy (pre-align and post-align). Specifically, we aligned molecular coordinates to a fixed orientation before or after training, rather than applying random rotations to the input molecular coordinates. The goal was to eliminate the impact of molecular rotations, and suit for tasks with strict requirements for rotational invariance. 
As shown in Table \ref{tab:3k}, pre-align refers to alignment performed before training, while post-align refers to alignment performed after training. We can observe that, except for a few tasks where the pre-align method exhibits marginal advantages, post-align performs better in most tasks. 
We attribute this performance degradation to the fact that the model can only learn 3D molecular information from a constrained, predefined perspective, which limits its generalization ability. On the other hand, random rotations preserve the rotational invariance of molecular geometry, allowing the model to focus on learning the intrinsic characteristics of molecules. Thus, our experimental results demonstrate that for small datasets, pre-aligning the coordinates beforehand may harm the model's fitting ability. 
The performance of the post-align method is similar to the method without alignment. Nevertheless, sometimes the alignment process may introduce slight shifts in the center point, leading to minor performance degradation. Moreover, post-align achieves rotational invariance and demonstrates overall better performance compared to manually extracted features.


\subsection{Performance on C10 Dataset}


The C10 Dataset comprises over 10k monoterpene carbocation intermediates with diverse 3D geometries and high-quality energy labels, which embody subtle conformation-dependent differences that are difficult to capture. These challenges are amplified by the fact that, despite sharing similar 2D structures, the 3D spatial arrangements and their associated energies vary considerably. 

As shown in Table \ref{tab:c10_mae}, through better encoding of the 3D information, our method leads to a lower MAE and RMSE, as well as an improved R$^2$. This indicates that it significantly enhances the predictive performance of the baseline model and demonstrates its potential in capturing complex molecular geometries, making it a promising candidate for tasks in quantum chemistry and related fields.

\begin{table}[H]
\centering
\caption{MAE, RMSE, R$^2$ Results on C10}
\begin{tabular}{cccc}
\toprule
Methods & MAE (kcal/mol) $\downarrow$ & RMSE (kcal/mol) $\downarrow$ & R$^2$ $\uparrow$ \\
\midrule
AttentiveFP\cite{xiong2019pushing} & 1.832 & 2.697 & 0.945  \\
+PointNet\cite{zhangPointGATQuantumChemical2024} & 1.616 & 2.490 & 0.950 \\
+Our Method & \textbf{1.523} & \textbf{2.423} & \textbf{0.951} \\
\bottomrule
\end{tabular}
\label{tab:c10_mae}
\end{table}


\subsection{Extended experiment}

\begin{figure}[ht]
    \centering
    \includegraphics[width=0.8\textwidth]{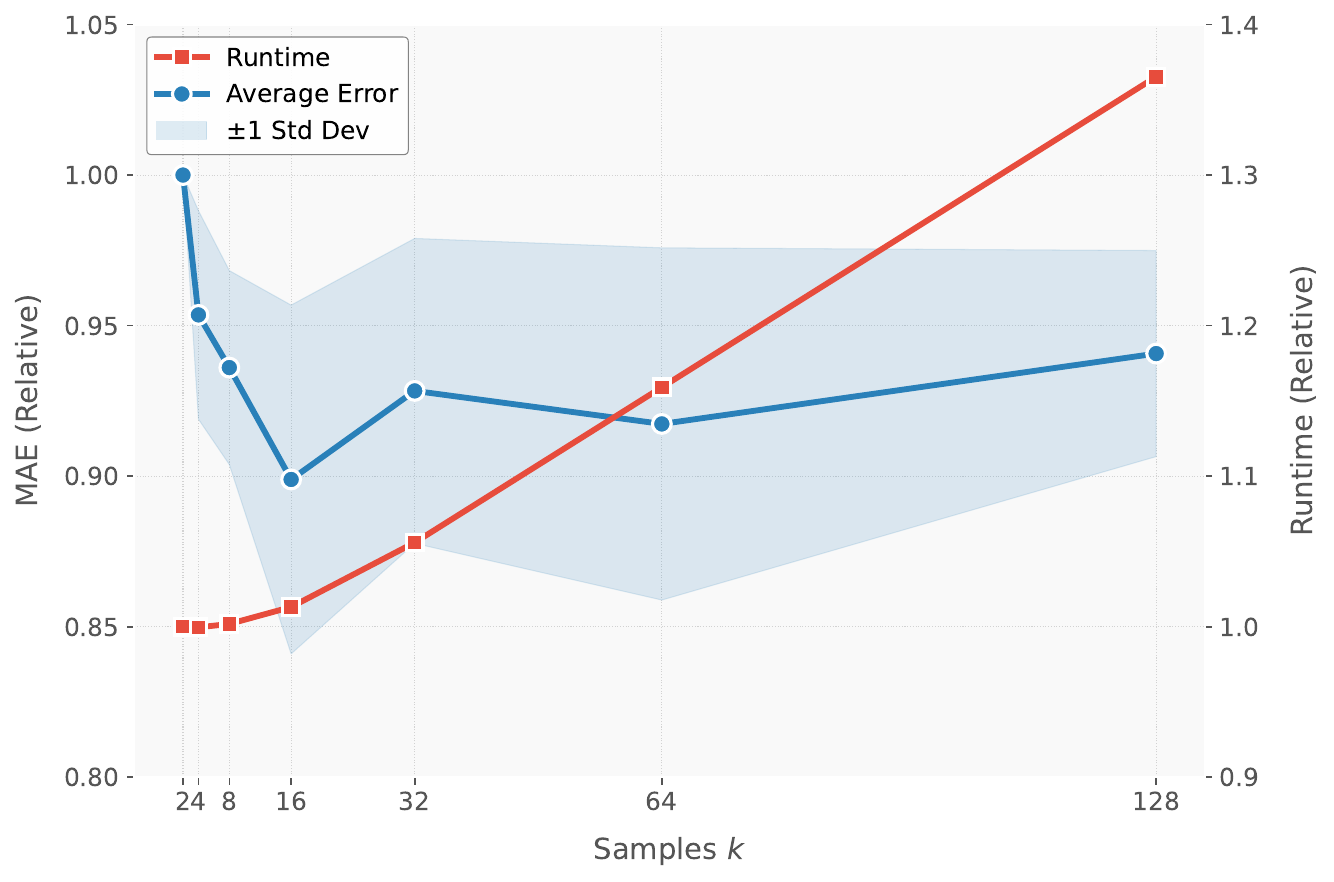} 
    \caption{Relative MAE Results and Runtime of different sample sizes $k$.}
    \label{fig:pio_vs_k}
\end{figure}

Fig.~\ref{fig:pio_vs_k} shows the overall performance of our method with different sample values $k$. It can be observed that when $k$ increases, the MAE for most properties initially decreases significantly and then starts to fluctuate. This indicates that more sampling angles provide more comprehensive geometric feature information, which helps enhance overall prediction performance. However, as the $k$ value continues to grow (e.g., $k=32$, $k=64$, $k=128$), most metrics begin to fluctuate around lower MAE values, demonstrating that most parameters do not require larger $k$ values.
Additionally, due to our model optimizations, the sampling for the model can be computed in parallel, which prevents significant differences in training computational cost (e.g., $k=2$ takes 27 minutes, $k=128$ takes 40 minutes). The choice of $k$ should depend on the specific task requirements. Here, we selected $k=16$.

\subsection{Ablation Studies}
The ablation study results, as shown in Table \ref{tab:abl_qm9_combined}, highlight several key insights into the performance of our model under different conditions. 
First, our model demonstrates strong performance on the smaller dataset, indicating its robustness and effectiveness in data-scarce scenarios. 
Second, removing our proposed method ("-Our Method") results in a substantial performance decline on both the  datasets, highlighting the central role of our method in capturing critical features. 
Third, engineered features are more important on the larger dataset, which is consistent with our previous observations. However, without the engineered features, our model still made a significant contribution.
Fourth, the removal of the PointNet module ("-PointNet") leads to the most significant performance degradation across all metrics and both datasets, which underscores the foundational role of PointNet in the architecture.
Overall, the performance drop is more pronounced on the smaller dataset, suggesting that our method is particularly effective at leveraging the additional data available in the data-scarce scenarios.

\begin{table}[ht]
    \centering
    \caption{Results of the ablation study across multiple metrics on QM9}
    \label{tab:abl_qm9_combined}
    \begin{tabular}{lSSSSSSS}
        \toprule
        \multirow{2}{*}{Metrics} & \multicolumn{3}{c}{QM9 (3k)} & \multicolumn{3}{c}{QM9 (10k)} \\
        \cmidrule(lr){2-4} \cmidrule(lr){5-7}
        & {-Features} & {-Ours} & {-PointNet} & {-Features} & {-Ours} & {-PointNet} \\
        \midrule
        A & -3.84\% & -17.67\% & -20.10\% & -13.60\% & -11.39\% & -17.24\% \\
        B & -2.04\% & -12.98\% & -35.18\% & -12.54\% & -6.39\% & -24.82\% \\
        C & -5.57\% & -9.58\% & -44.83\% & -13.95\% & -3.27\% & -36.16\% \\
        Cv & -18.80\% & -18.14\% & -43.87\% & -24.08\% & -15.47\% & -32.15\% \\
        G$_\text{atom}$ & -13.76\% & -22.51\% & -50.60\% & -28.07\% & -17.28\% & -37.53\% \\
        G & -18.06\% & -18.89\% & -47.30\% & -32.43\% & -21.86\% & -34.64\% \\
        H$_\text{atom}$ & -14.10\% & -23.02\% & -51.26\% & -28.97\% & -17.95\% & -38.02\% \\
        H & -17.69\% & -19.26\% & -47.09\% & -32.41\% & -21.92\% & -34.61\% \\
        U0$_\text{atom}$ & -14.16\% & -22.67\% & -51.26\% & -28.85\% & -17.91\% & -38.00\% \\
        U0 & -17.90\% & -19.06\% & -47.32\% & -32.45\% & -21.82\% & -34.66\% \\
        U$_\text{atom}$ & -14.47\% & -22.36\% & -51.66\% & -28.93\% & -17.89\% & -37.99\% \\
        U & -17.59\% & -19.45\% & -46.99\% & -32.42\% & -21.91\% & -34.61\% \\
        alpha & -6.71\% & -16.49\% & -48.66\% & -12.85\% & -15.96\% & -33.72\% \\
        gap & -6.49\% & -6.55\% & -12.99\% & -9.67\% & -1.23\% & -9.71\% \\
        homo & -3.52\% & -6.11\% & -12.27\% & -9.22\% & -1.43\% & -7.45\% \\
        lumo & -9.31\% & -1.75\% & -17.63\% & -9.28\% & -2.57\% & -11.14\% \\
        mu & -0.83\% & -2.73\% & -7.24\% & -1.33\% & -0.99\% & -4.85\% \\
        r2 & -3.30\% & -14.92\% & -50.40\% & -23.79\% & -7.83\% & -36.75\% \\
        zpve & -1.56\% & -15.29\% & -41.58\% & -15.62\% & -5.74\% & -28.37\% \\
        \bottomrule
    \end{tabular}
\end{table}

\section{Discussion}

\subsection{Learned Representations vs. Engineered Features on Small Datasets}

The results on the small dataset (QM9 3k and 10k) indicate that our method for learning 3D representations outperforms approaches relying on engineered features (e.g., distances and angles). While engineered features can encapsulate certain structural and chemical information, their inherent limitations in expressiveness make them less flexible in scenarios that require detailed 3D information. In contrast, the adaptive learning process of our method enables it to capture subtle 3D characteristics without being constrained by manually engineered priors. The further performance improvement observed when combining our method with these features highlights the potential of a hybrid strategy that leverages both data-driven and domain-specific insights.

\subsection{Rotational Invariance and Alignment Strategies}

We evaluated the impact of fixed molecular alignment on model performance. The results show that enforcing pre-alignment on the dataset has a negative effect, potentially causing the model to learn noise and limiting its generalization capability. Experimental comparisons between pre-alignment, post-alignment, and no alignment demonstrate that allowing the model to independently learn rotational invariance (no alignment) achieves the best performance. Post-alignment also maintains comparable performance. However, due to changes introduced during the alignment process, particularly in determining the center point, noise may be introduced, which slightly impacts performance. From the perspective of rotational invariance, our method calculates the expectation under SO(3), which not only enhances performance but also improves the stability of rotational invariance. For tasks with strict rotational invariance requirements, post-alignment serves as an excellent alternative.

\subsection{Interpretability}
\begin{figure}[H]
    \centering
    \includegraphics[width=0.9\textwidth,trim={2cm 2cm 2cm 2cm},clip]{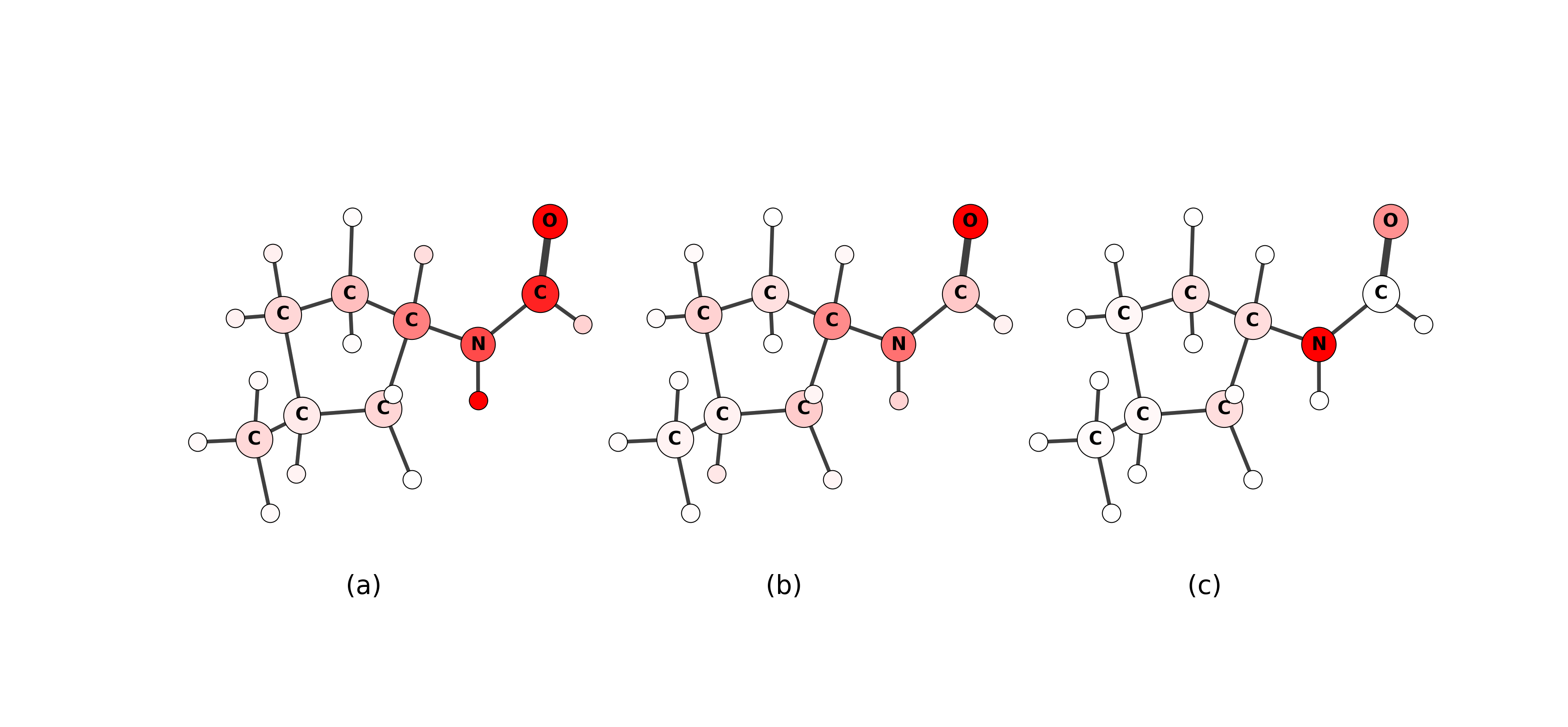} 
    \caption{Atom importance visualization for cyclopentane derivative: (a) Engineered feature method method highlighting C atoms in the amide group; (b) Our method showing importance distribution around the N and O atom; (c) Calculation result focus mostly on the N and O atom}
    \label{fig:m19}
\end{figure}

\begin{figure}[H]
    \centering
    \includegraphics[width=0.9\textwidth,trim={2cm 2cm 2cm 2cm},clip]{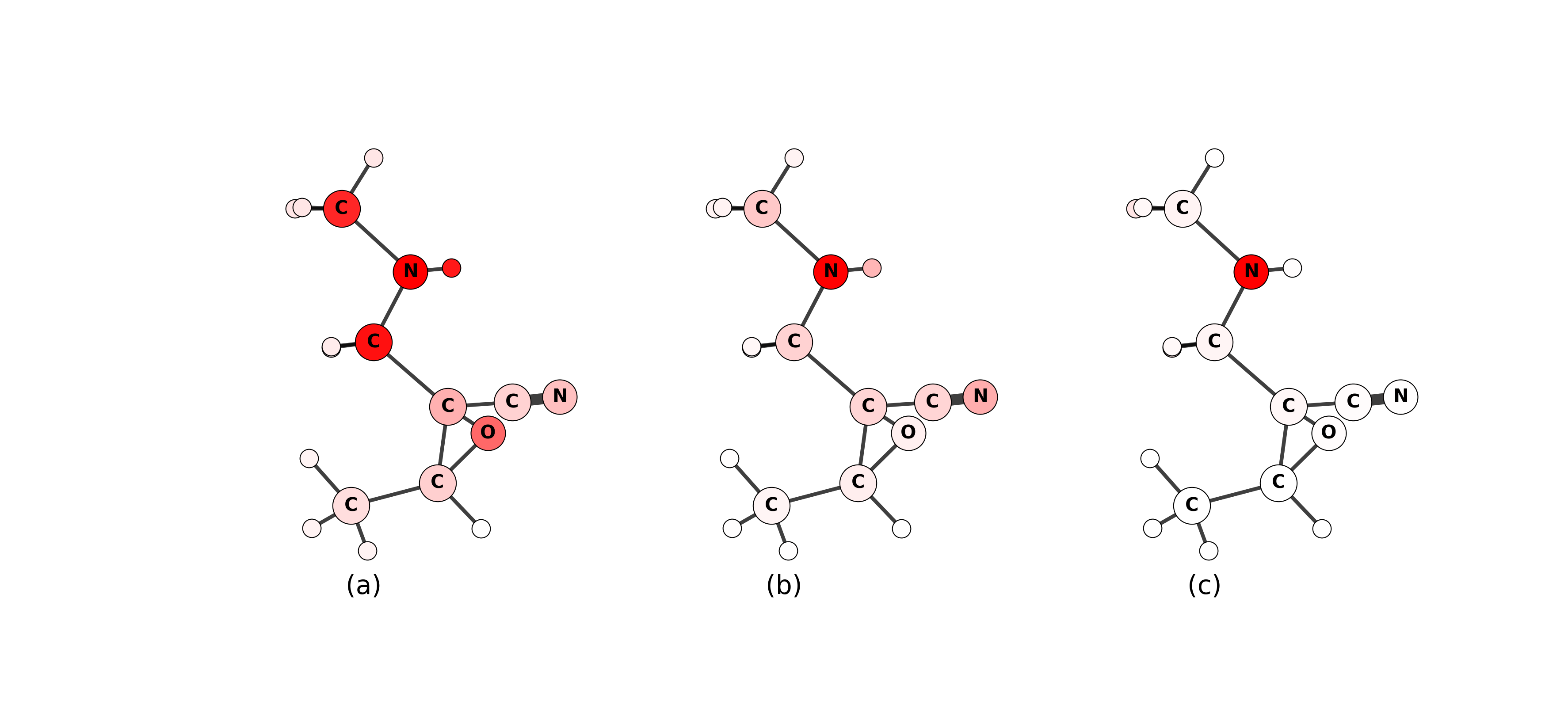} 
    \caption{Atom importance visualization: (a) Engineered feature method emphasizing the amino part; (b) Our method focus more on the N atom; (c) Calculation result focus most on the N atom}
    \label{fig:m50}
\end{figure}

\begin{figure}[H]
    \centering
    \includegraphics[width=0.9\textwidth,trim={2cm 2cm 2cm 3cm},clip]{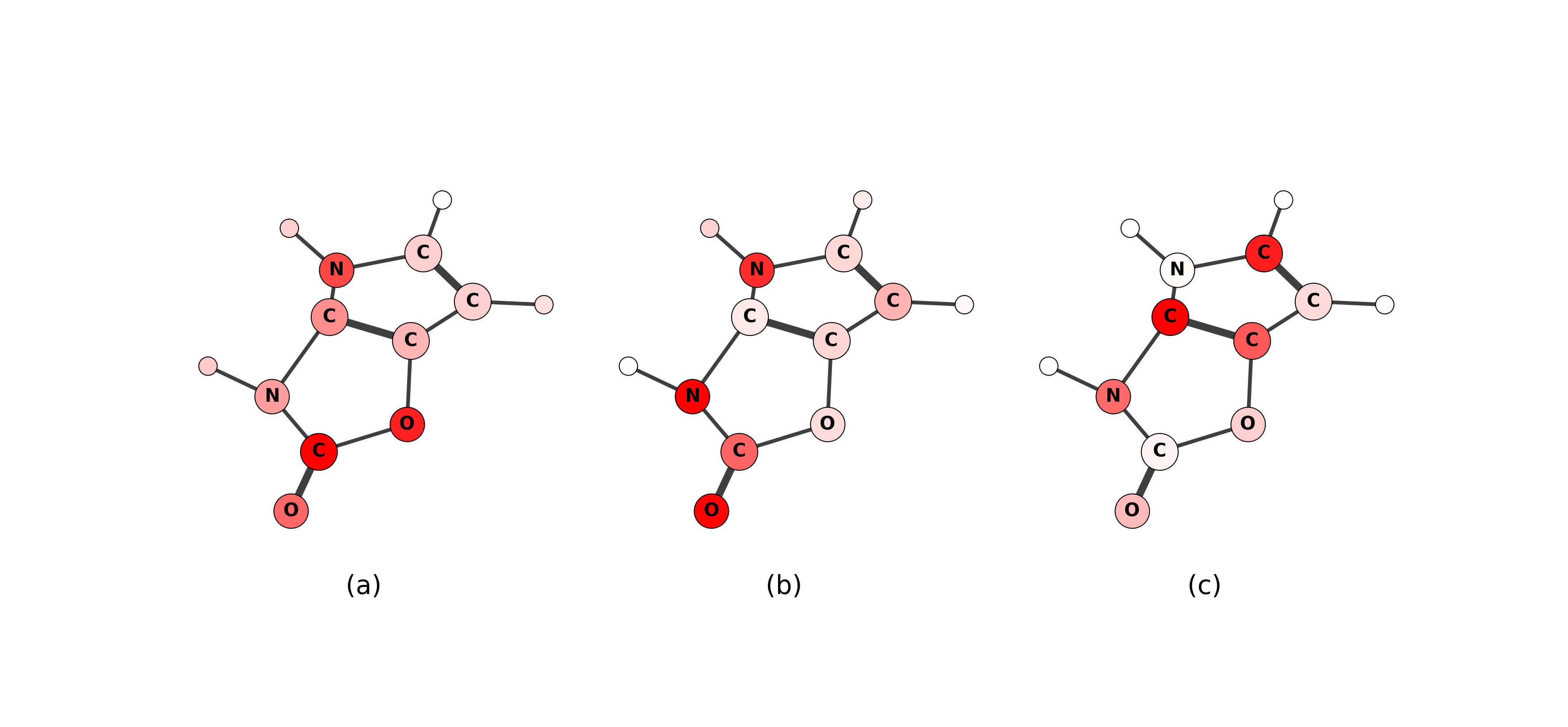} 
    \caption{Atom importance visualization: (a) and (b) GNN based method showing a different pattern from (c) calculation result.}
    \label{fig:m14}
\end{figure}

Feature importance assessment plays an important role in understanding and interpreting model behavior. Here, we calculated the importance of atom contributions to \textit{homo}, as shown in Fig.~\ref{fig:m19}, \ref{fig:m50}, and \ref{fig:m14} through: (a) Engineered feature method and (b) our method computing gradients of the output with respect to input features, and (c) calculations based on PySCF.

As shown in Fig.~\ref{fig:m19} and Fig.~\ref{fig:m50}, our method produces importance distributions that more closely align with the computed ground truth than conventional approaches. This improvement is particularly evident in regions of high feature significance, where our method more accurately identifies and weights critical elements of the input data. The enhanced spatial encoding mechanism appears to capture structural dependencies that standard GNNs may overlook, resulting in more faithful importance attribution. 
We randomly selected 20 samples to calculate the average error, and the results show that our method is closer to the calculated results (RMSE: 0.3331 vs. 0.2917, MAE: 0.2229 vs. 0.1875).

However, we observe that for certain samples, GNN methods exhibit substantial discrepancies from the computational result, as shown in Fig.~\ref{fig:m14}. GNN approaches may be influenced by the inherent message-passing and neighborhood aggregation mechanisms of graph neural networks. This structural bias sometimes leads GNNs to emphasize features different from the theoretical optimum, particularly at graph boundaries or in regions with complex connectivity patterns. These differences highlight the limitations of MPNN.

\section{Conclusion}

This paper proposes a method to encode 3D information in the field of molecule properties prediction. 
This method offers several advantages: First, it requires no prior knowledge, which enables strong generalization performance across diverse datasets and tasks. Second, it balances feature extraction capability with rotational invariance, ensuring robust performance on 3D molecular data. Third, the method is designed as a decoupled module, making it a plug-and-play solution that can be easily integrated into existing frameworks. Finally, it supports parallel computation without the need for $n$-hop message passing, significantly improving computational efficiency.

Despite its advantages, the proposed method has some limitations. First, there exists an inherent error in achieving perfect rotational invariance, with the error magnitude approximately scaling as $\mathcal{O}\left(\frac{1}{\sqrt{k}}\right)$. Second, the aggregation method used in the model, such as \textit{mean}, is less robust compared to \textit{max} aggregation. However, addressing this limitation is challenging, even for approximate solutions.

Future research could focus on several directions. One potential avenue is to further explore the integration of engineered features with automatically extracted features, aiming to optimize the balance between predefined domain knowledge and adaptive feature learning. Additionally, while current studies using PointNet for molecular information extraction rely on global embeddings, constructing local embeddings could potentially enhance model performance. Moreover, the obtained 3D embeddings currently lack edge information, and investigating methods to incorporate edge information could be another promising direction for future work.

\bibliographystyle{ieeetr}
\bibliography{lib}

\end{document}